\documentclass[runningheads]{llncs}

\usepackage[T1]{fontenc}
\usepackage{microtype}
\usepackage{hyperref}
\usepackage{enumitem}
\usepackage{orcidlink}

\renewcommand{\orcidID}[1]{\unskip$^{\,\mbox{\scriptsize\orcidlink{#1}}}$}

\hypersetup{
  pdftitle={Scaling Accessible Mathematics on arXiv: HTML Conversion and MathML 4},
  pdfauthor={Deyan Ginev, Bruce R. Miller, Brian Caruso, Jacob Weiskoff, Jeff Sank},
  colorlinks=true,
  linkcolor=blue,
  urlcolor=blue,
  citecolor=blue
}

\begin{document}

\title{Scaling Accessible Mathematics on arXiv:\\HTML Conversion and MathML~4}
\titlerunning{Scaling Accessible Mathematics on arXiv}
\author{Deyan Ginev\inst{1}\orcidID{0000-0002-7306-8634} \and
Bruce R. Miller\inst{2}\orcidID{0000-0002-2300-0367} \and
Brian Caruso\inst{1}\orcidID{0000-0002-7489-0144} \and
Jacob Weiskoff\inst{1}\orcidID{0000-0002-8778-0032} \and~Jeff~Sank\inst{1}}
\authorrunning{D. Ginev et al.}
\institute{arXiv, Cornell Tech, New York NY, USA\\
\email{\{deyan,bcaruso,jake,jeff\}@arxiv.org}
\and
National Institute of Standards and Technology, Gaithersburg MD, USA\\
\email{bruce.miller@nist.gov}}

\maketitle

\begin{abstract}
We report on the ongoing development of arXiv's HTML Papers offering, available on every
new TeX/LaTeX submission since its initial release in 2023. The main highlights
from 2025 and early 2026 are:
\begin{itemize}[noitemsep, topsep=0pt, parsep=2pt, partopsep=0pt]
\item[(i)] community-driven improvements to HTML fidelity and service health,
with roughly half of 6{,}000 user reports resolved;
\item[(ii)] corpus-scale conversion work aimed at 90\% error-free HTML\newline (currently 75\%);
\item[(iii)] initial MathML~4 Intent annotations for accessible speech output;
\item[(iv)] an in-progress Rust port of LaTeXML, reducing compute costs
and enabling faster previews on submission.
\end{itemize}
The arXiv HTML Papers project remains experimental, but is gradually maturing
as we better understand the needs of arXiv's readers and the technical
opportunities presented by new standards and by advances in programming languages and AI.

\end{abstract}

\section{Background}
arXiv is the world's largest preprint server, covering mathematics, computer science, physics,
and related STEM fields. Over 3 million articles have been published to date, and growth is
still accelerating: monthly submissions reached 30{,}000 for the first time in
2026~\cite{arxiv-monthly-submissions}. Its dominant distribution format, PDF, preserves visual
fidelity but offers limited structural information for reflow, interaction, and assistive technologies. Ongoing work by the LaTeX team aims to improve PDF accessibility~\cite{mittelbach2025mathml}.
HTML with high-quality MathML is a different kind of representation: it exposes narrative
and mathematical structure to the Web Platform~\cite{web-platform}, where the same document asset
can simultaneously support responsive rendering in a browser, accessible readouts via screen readers,
contextual navigation, and machine-readability for search, linking, and downstream analysis.

A pipeline at arXiv scale must confront the full diversity of STEM documents: heterogeneous notations,
varied narrative statements, visual blocks (figures, tables, diagrams), rich metadata and backmatter,
as well as idiomatic TeX across every discipline. Our conversion process offers a contract: once
a LaTeX document is successfully processed by our system, the authored content survives intact to the
final HTML, and every structural or semantic LaTeX construct becomes HTML markup that the Web Platform can act on.
Success has degrees: simple sectioning is easy, responsive design and theming are relatively
low effort, accessible readouts of mathematical expressions are harder, and downstream affordances
such as rich, accessible navigation of commutative diagrams are harder still.

Reaching the threshold where 90\% of arXiv submissions receive successful HTML conversion will be
a strong signal that LaTeX-authored documents can be made accessible via HTML at scale, and that
arXiv itself can provide an official service of this nature long-term. Guaranteeing high-quality
accessible outcomes for a given document is, however, only possible when ``the authors have a working
understanding of accessibility, and write with accessibility in mind''~\cite{ross2025accessibility,arxiv-latex-best-practices}.

\section{MathML Intent and accessibility}

Our HTML output carries MathML rather than formula images. That choice exposes mathematical structure
 to the Web Platform (navigable expressions, speech output, responsive reflow) and lets us
 incrementally enrich the same markup as browser support and accessibility tooling mature. 
 This direction was enabled in 2023, when Igalia's work brought MathML Core~\cite{mathml-core} to
 Chromium~\cite{igalia-mathml-core-2023} and completed cross-browser support: native mathematics 
 rendering became viable in every major browser for the first time. LaTeXML and ar5iv had initially 
 demonstrated that a baseline HTML+MathML conversion is viable at arXiv scale~\cite{latexml,ar5iv}; 
 the current challenge is to build on that work and create high-quality readouts for the full range 
 of documents and expressions in arXiv's own pages.

arXiv hosts cutting-edge STEM research, where notation is often specialised, locally overloaded,
or introduced for the first time. In this setting, default assistive technology (AT) readouts can be unreliable or confusing
for listeners, and it remains an open research question how far remediation of the most novel
and most difficult notations can be taken.
The relevant standard for addressing this is MathML~4 Intent~\cite{mathml4-intent}. The new \texttt{intent} attribute carries
 a compact expression syntax and standardises three vocabularies of reserved values: \emph{core concepts},
\emph{open concepts}, and \emph{core properties}. It also defines an underscore syntax for
\emph{literals}: strings meant to be pronounced as-is by AT.
The Core vocabulary targets secondary and post-secondary education content and offers standard solutions for the
most common problems in STEM texts.

Full remediation of novel notations is not our initial focus. Our current
baseline is syntax-first. We annotate formulae with the \texttt{:literal} intent property, 
which is trivial to assign systematically and gives a predictable target for speech output. 
Alongside \texttt{:literal} we emit well-structured, navigable MathML Core, constructed from the
author's TeX via LaTeXML's math grammar. We expect that LaTeXML's rich internal representation of mathematical expressions can serve as a
fruitful base on which to build MathML trees using Intent's Open concepts.

Our MathML output also carries the original TeX source for each formula as an attached annotation 
inside the \texttt{<math>} element. This lets non-standard services that rely on TeX syntax operate
directly on the same HTML page without further modification. We caution, however, that the well-behaved 
dialect of TeX that is common in web-native authoring is a small subset of the TeX used in arXiv 
mathematics. Consumers targeting the full arXiv corpus should prefer the MathML Core trees,
whose restricted vocabularies of elements and attributes, together with a standard tree structure,
yield predictable downstream applications and behaviour.

\section{arXiv deployment and new HTML page scaffold}

arXiv's HTML Papers offering went live in 2023~\cite{Frankston2024HTMLPO}, surfacing 
an experimental HTML rendition on every \emph{new} submission for which LaTeXML produces output. 
The historical arXiv corpus is not yet part of the live pipeline and is currently available in HTML 
form only at the ar5iv Labs site~\cite{ar5iv}; bringing historical articles into the main HTML Papers
channel is a future goal. The live pipeline is updated on a roughly quarterly cadence. Each update 
folds in kernel improvements from upstream LaTeXML at NIST~\cite{latexml} together with 
community-contributed enhancements and arXiv's own robustness patches. Every LaTeXML fix is 
contributed upstream first, as a pull request to the NIST-maintained repository; it is then either 
merged at NIST, or applied to arXiv's ``high-velocity'' 
LaTeXML fork during a preview window. Our shared goal is for every upgrade used by arXiv to land upstream,
so that the two codebases stay in close alignment.

In parallel with the conversion pipeline, we iterated on the page scaffolds that wrap 
each converted article, deploying a new version in March 2026. 
The rework cleanly separates page markup, theme, and in-page services, so each can evolve independently.
This allowed us to address common reader requests for a less obtrusive reading experience, such as
options to minimise the surrounding UI and better handling of oversized constructs on narrow devices.

\section{Impact signals: user reports and corpus statistics}

We prioritise work using two complementary signals that surface different failure modes.

The first is the public \texttt{arXiv/html\_feedback} issue tracker on GitHub, where readers and 
authors report conversion problems on individual articles~\cite{html-feedback}. These reports are
a direct view into the \emph{most acute} errors: the rendering defects that motivated a reader to
open an issue. From the start of 2025 through April 2026 we have resolved half of 
approximately 6{,}000 reports received.

The second is large-scale missing-package statistics gathered by running the LaTeXML pipeline across
the historical ar5iv corpus~\cite{ar5iv}, which covers the roughly 90\% of arXiv articles that have
TeX/LaTeX source (the remaining 10\% are PDF-only). This provides an aggregate view of which 
LaTeX packages and macros most frequently cause conversion to fail, and identifies the 
\emph{most frequent} errors.

Optimising only for frequency can leave acutely broken papers in place, while optimising only for
reported friction can miss subtle but pervasive errors; tracking both keeps our work scalable and
visible to readers.

\section{Corpus coverage and LaTeXML v0.9}

We can report two distinct coverage metrics for the arXiv HTML output. About 97\% of submissions 
currently produce some HTML; that figure is an availability ceiling rather than a quality claim, 
and includes documents that render partially or badly. The more meaningful metric is the fraction 
of articles that convert without LaTeXML errors and are typically mostly readable. That number
has slipped to roughly 75\%. The slippage tracks a moving target in
the literature itself: conferences, journals, and authors adopt new macro packages, and dependencies
continue to migrate toward newer LaTeX 3 definitions.

The largest structural change to LaTeXML in this cycle is an improved kernel for raw LaTeX 3 interpretation.
Now arXiv exercises that kernel fully: packages without LaTeXML bindings are loaded raw rather
than skipped, falling back to a simpler HTML dialect only when their semantics are not recoverable
(with corresponding trade-offs in accessibility, themability, and responsive layout). Alongside
the kernel work, sustained improvements have landed in the vector-graphics and math-rendering
paths most common in arXiv preprints, leading to broader coverage of TikZ and xy diagrams.

The upcoming LaTeXML~v0.9 also delivers a rework of frontmatter handling: titles, authors,
affiliations, and associated metadata now flow through a more uniform model, with many class
bindings updated to match. Together with the kernel and graphics work above, this directly
benefits arXiv, whose papers routinely carry intricate author metadata.

NIST's Perl LaTeXML remains the production reference. Our aim is to fold successful advances from 
the arXiv preview channel into the reference repository in time for the LaTeXML v0.9 release, 
expected in summer 2026.

\section{A Rust reimplementation of LaTeXML}

We report on our ongoing Rust reimplementation of LaTeXML, started at NIST and continuing at arXiv.
It is driven by three concerns: reducing cloud conversion costs at arXiv scale; improving the submitter
preview experience through faster conversion; and sustainability~--~arXiv is itself migrating away from
Perl, which is no longer a mainstream choice for new projects and whose contributor community continues to shrink.

Work began in 2016, with renewed focus between 2020 and 2024, producing roughly 50{,}000 lines of
hand-written Rust covering the TeX-engine emulation, tokenization, digestion, and XML model. This was enough
to pass about 25\% of the core LaTeXML test suite.

The translation effort had remained largely dormant until the arrival of Claude Opus~4.6 and 4.7
in February and April 2026, whose agentic coding loop made sustained progress practical. In roughly
three weeks of AI-assisted development we advanced what we estimate would otherwise have been more
than two person-years of equivalent hand-written work. The agentic contribution is now approaching
100{,}000 additional lines of Rust, and the pipeline passes all core tests. We have also reached
output parity with the Perl implementation on the first few hundred arXiv articles in our evaluation slice.

We use ``parity'' in an engineering sense: the Rust pipeline succeeds on the same inputs, preserves essential document
structure, and produces sufficiently similar HTML/MathML for downstream accessibility and rendering.
Early benchmarks on matched inputs show the Rust pipeline running 10--30 times faster than the Perl
reference, supporting the cost and preview-latency goals that motivated the translation.
The code is presently a private repository; in keeping with the licensing stance of upstream LaTeXML,
it is intended for permissive public-domain release as soon as we have confidence in the parity of the translation.
This preserves continuity: LaTeXML's broad user base should continue to use the production-ready
Perl original until we are confident that a switchover is viable.

We now regard the translation effort as a case study in when agentic AI can accelerate the
modernisation of mature scientific software. So far we have identified seven guardrails that lead to
a successful initial phase of agentic translation:
\begin{enumerate}
  \item a bounded translation task (from a feature-complete Perl design to Rust);
  \item a substantial hand-written bootstrap preceding any AI use;
  \item an executable test suite of 332 cases spanning the LaTeXML pipeline;
  \item immediate feedback from \texttt{rustc}, \texttt{clippy}, and a RelaxNG schema for structural output validation;
  \item access to canonical documentation (the LaTeXML manual, Knuth's \emph{TeXbook}, and the TeX source code itself);
  \item access to the upstream LaTeXML git history as a contextual guide at PR granularity;
  \item the model's cross-language expertise across TeX, Perl, Rust, Unicode, XML, and HTML.
\end{enumerate}

There is also one backstop: roughly hourly review by a senior maintainer, who catches the
residual failure modes that survive the seven guardrails.

We present the list less as a claim about any particular generation of models than as a recipe
for modernising mature scientific software: rely on a test suite and compiler feedback, commit to
a well-scoped translation, provide initial guidance with hand-written code,
and keep a senior maintainer in the loop.

\section{Outlook}

The long-term goal is to make born-accessible mathematical communication the default rather than the
exception. arXiv is a natural setting for that shift: an open, live, continuously growing corpus where
improvements propagate across a major portion of the mathematical literature at once.

arXiv is also at a moment of institutional transition. In April 2026, arXiv announced 
that it is leaving its long-standing home at Cornell Tech and becoming an independent
nonprofit~\cite{arxiv-independent-2026}. The new structure aims to give arXiv a renewed stability
and focus, at the scale the community now expects.

We welcome feedback from the ICMS community: ideas for collaboration on better HTML Papers
and accessible mathematics, as well as general thoughts on how you'd like to use arXiv in the future.

\begin{credits}
\subsubsection{Author contributions.} DG led this work and is
sole developer of the Rust LaTeXML reimplementation. BRM contributes upstream LaTeXML and the v0.9
release path at NIST. BC, arXiv's technical lead, supervised user-facing architecture
and production deployments. JW and JS provide institutional leadership at arXiv, with JW additionally
serving as senior expert on the TeX/LaTeX use of arXiv's authors.
\end{credits}

\bibliographystyle{splncs04}
\bibliography{refs}

\end{document}